\documentclass[11pt]{article}
\usepackage{graphicx}
\usepackage{amssymb}
\usepackage{amscd}
\usepackage{mathrsfs}
\usepackage{longtable,lscape}
\usepackage{amsthm}
\usepackage{amsfonts}
\usepackage{amsmath}
\usepackage{bbm}
\usepackage{float}
\usepackage{url}

\newcommand{\captionfonts}{\footnotesize}
\makeatletter  % Allow the use of @ in command names
\long\def\@makecaption#1#2{%
  \vskip\abovecaptionskip
  \sbox\@tempboxa{{\captionfonts #1: #2}}%
  \ifdim \wd\@tempboxa >\hsize
    {\captionfonts #1: #2\par}
  \else
    \hbox to\hsize{\hfil\box\@tempboxa\hfil}%
  \fi
  \vskip\belowcaptionskip}
\makeatother 
\begin{document}
\title{Quantum-theoretic Modeling in Computer Science \\
\Large A complex Hilbert space model for entangled concepts \\ in corpuses of documents\footnote{Submitted to: {\it International Journal of Theoretical Physics}}}
\author{Diederik Aerts$^1$, Lester Beltran$^1$, Suzette Geriente$^1$ and Sandro Sozzo$^2$
 \vspace{0.5 cm} \\ 
        \normalsize\itshape
        $^1$ Center Leo Apostel for Interdisciplinary Studies, 
         Brussels Free University \\ 
        \normalsize\itshape
         Krijgskundestraat 33, 1160 Brussels, Belgium \\
        \normalsize
        E-Mails: \url{diraerts@vub.ac.be,lbeltran@vub.ac.be} \\
        \url{sgeriente@vub.ac.be}
          \vspace{0.5 cm} \\ 
        \normalsize\itshape
        $^2$ School of Business and Centre IQSCS, University of Leicester \\ 
        \normalsize\itshape
         University Road, LE1 7RH Leicester, United Kingdom \\
        \normalsize
        E-Mail: \url{ss831@le.ac.uk} 
            	\\
              }
\date{}
\maketitle
\begin{abstract}
\noindent
We work out a quantum-theoretic model in complex Hilbert space of a recently performed test on co-occurrencies of two concepts and their combination in retrieval processes on specific corpuses of documents.  The test violated the Clauser-Horne-Shimony-Holt version of the Bell inequalities (`CHSH inequality'), thus indicating the presence of entanglement between the combined concepts. We make use of a recently elaborated `entanglement scheme' and represent the collected data in the tensor product of Hilbert spaces of the individual concepts, showing that the identified violation is due to the occurrence of a strong form of entanglement, involving both states and measurements and reflecting the meaning connection between the component concepts.  These results provide a significant confirmation of the presence of quantum structures in corpuses of documents, like it is the case for the entanglement identified in human cognition.
\end{abstract}
\medskip
{\bf Keywords}: CHSH inequality, Quantum entanglement, Quantum structures, Corpuses of documents

\section{Introduction\label{intro}}
In the last decade, the `quantum cognition research programme' has successfully confirmed the presence of genuine quantum structures in complex cognitive processes, like conceptual categorization, probability judgement and decision-making
\cite{aerts2009,khrennikov2010,busemeyerbruza2012,aertsbroekaertgaborasozzo2013,aertsgaborasozzo2013,kwampleskacbusemeyer2015,dallachiaragiuntininegri2015a,dallachiaragiuntininegri2015b,haven2018}
and the corresponding applications in linguistics and information retrieval 
\cite{schmittetal2008,bruza2009,coecke2010,piwowarskietal2010,frommholzetal2010,Ingo2011,bucciomeluccisong2011,melucci2015,aertsIJTP,aertsetal2018e}.

In the present paper we focus on the identification and modeling of undoubtedly one of the most fascinating and important quantum phenomena called `entanglement' in corpuses of documents. 
Historically, the notion `entanglement' was introduced by Erwin Schr\"odinger in a letter to Albert Einstein, after reading the so called `Einstein Podolsky Rosen paradox' article where Einstein together with two of his collaborators 
Boris Podolsky and Nathan Rosen analysed a specific type of correlations, now commonly called `EPR-correlations', appearing in a compound quantum entity consisting of two sub entities \cite{einstein1935}. 
In the 
 same year, Schr\"odinger published an article on the matter \cite{schrodinger1935}, where he officially %SS introduces 
 introduced 
 the notion of `entanglement', and like the title indicates, entanglement was imagined to appear for two `separated (in space) quantum entities', while the phenomenon seems to indicate that the entities are `not separated', which made Einstein use the expression `spooky action at a distance' when talking about the phenomenon. 
 Later, 
 David Bohm introduced the 
nowadays 
 archetypical example of the EPR-type of correlation and entanglement of the two spin 1/2 particles in a singlet spin state of zero total spin and flying apart while quantum theory predicts the spin correlations to persist \cite{bohm1951}. If one imagines that this means that if one spin is forced ``up'' by the measuring apparatus applied to it then, as a consequence `immediately' the other spin is forced ``down'', even when not measured upon, and this type of `mechanism' keeps taking place for two quantum particles independent of the distance in space they are apart, the phenomenon is indeed a `spooky action at a distance'. It was John Bell who introduced a way to experimentally testing for the presence of entanglement by formulating an inequality, meanwhile called Bell's inequality, which when violated proved the presence of entanglement \cite{bell1964,bell1987}, and it is the the Clauser Horne Shimony Holt (CHSH) version of Bell's inequality \cite{clauser1969} which we will consider in the present paper. The CHSH inequality was also the one considered by Alain Aspect when he showed it to be violated experimentally by the entangled spin quantum entity \cite{aspect1982}, and meanwhile many experiments have identified the presence of quantum entanglement. The phenomenon is also increasingly investigated with technical applications in mind, e.g. recently a new quantum entanglement distance record of 1203 km was realized, from the Micius satellite to bases in Lijian, Yunnan and Delingha, Quinhai, increasing the efficiency of transmission over prior fiberoptic experiments by an order of magnitude \cite{juanyin2017}.

Entanglement has also been investigated within quantum cognition
\cite{bruzakittonelsonmcevoy2009,aertssozzo2011,aertssozzo2014,bruzakittorammsitbon2015,gronchistrambini2017,aertsetal2018a,aertsetal2018b}. In this regard, in 2011 we designed a 
`composite entity consisting of sub entities setting' using the conceptual combination {\it The Animal Acts}\footnote{For {\it Acts} we have mainly considered `the action of emitting a sound'.} as a composite entity made up of the individual concepts {\it Animal} and {\it Acts}, and performed a cognitive test which significantly violated the CHSH inequality \cite{aertssozzo2011}. More recently, we considered the same conceptual combination and tested the CHSH inequality in various document retrieval searches on the web using specific corpuses of documents, showing that coincidence measurements on {\it The Animal Acts} violate the CHSH inequality \cite{beltrangeriente2018}. We precisely used three corpuses of documents in our investigation, the corpus `Google Books', the corpus `News on Web' (NOW) and the `Corpus of Contemporary American English' (COCA). Google Books is the biggest available corpus, with 560 billion words of books ranging over centuries and scanned by Google. Then comes the NOW corpus which 6 billion words of texts from news and periodicals, and finally 
COCA has 560 million words of texts of the types of stories.\footnote{The three corpuses are freely available on the web. More precisely, `Google Books' can be found at the website \url{https://googlebooks.byu.edu/x.asp}, NOW can be found at the website \url{https://corpus.byu.edu/now/}, and COCA can be found at the website \url{https://corpus.byu.edu/coca/}.} In all search operations we made, we found a significant violation of the CHSH inequality by an amount greater than the one we found in the cognitive test, and the amount of violation even exceeds the so-called `Cirel'son bound' for  coincidence measurements on quantum entities \cite{cirelson1980}. We also performed a variant of the operation applying to the corpus COCA the `collocates technique', which more closely resembles the combination operations of the human mind, and again found a violation of the CHSH inequality, but within the Cirel'son bound this time \cite{beltrangeriente2018}.

In this paper, we work out a quantum-theoretic modeling of the two web tests, the `Google Books test' and the `Collocates test', employing an entanglement scheme put forward in \cite{aertssozzo2014} and analysed in detail in \cite{aertsetal2019}. 
In line with our quantum cognition approach to model concepts and their combinations \cite{aerts2009,aertsbroekaertgaborasozzo2013,aertsgaborasozzo2013}, and taking into account our realistic-operational description of conceptual entities \cite{aertssassolidebianchisozzo2016}, we consider the conceptual combination {\it The Animal Acts} as a composite entity made up of the individual entities {\it Animal} and {\it Acts}, and the coincidence measurements as operations changing the states of the entities which they are performed on. There is however a specific aspect about the mathematical description and identification of `entanglement', which we identified in  \cite{aertssozzo2014}, and which is not well-known at all, and plays an essential role in the modeling we will work out. Let us put forward this aspect after having become somewhat more specific in our modeling scheme, because it is related to the use of the mathematical structure of complex Hilbert space in quantum theory. The coincidence measurements that we will consider on the conceptual entity {\it The Animal Acts} are experiments with 4 outcomes, and since they will be modeled by self-adjoint operators, the complex Hilbert space where this takes place will be $\mathbb{C}^{4}$ of all 4-tuples of complex numbers. The sub measurements on the two sub entities {\it Animal} and {\it Acts} of which the coincidence measurements on the composed entity {\it The Animal Acts} are compositions will be measurements with 2 outcomes, which means that self-adjoint operators on $\mathbb{C}^{2}$ the complex Hilbert space of all 2-tuples of complex numbers, seems at first sight the candidate to model these sub measurements. Indeed, it is a customary practice to use the tensor product $\mathbb{C}^{2} \otimes \mathbb{C}^{2}$ which is isomorphic as a complex Hilbert space with $\mathbb{C}^{4}$ and to model the coincidence measurements as products of two self-adjoint operators in $\mathbb{C}^{2}$ within this tensor product procedure. However, we know that exactly when `entanglement' is present in the considered situation the state of the composed entity represented by a unit vector of $\mathbb{C}^{4}$, will in general not be the product of two unit vectors of $\mathbb{C}^{2}$, indeed, it is exactly the crucial property of the tensor product $\mathbb{C}^{2} \otimes \mathbb{C}^{2}$ to contain such non product vectors, and that `are' the mathematical representations of the entangled states of the composed entity. However, we should not forget that also `entangled measurements' exist, and more precisely the vectors space $L(\mathbb{C}^{4})$ of all linear operators on $\mathbb{C}^{4}$ is also mathematically isomorphic to the tensor product $L(\mathbb{C}^{2}) \otimes L(\mathbb{C}^{2})$, where $L(\mathbb{C}^{2})$ of all linear operators on $\mathbb{C}^{2}$. A natural question arises hence, why would entanglement only appear in the states of a composed entity and not in the coincidence measurements that are performed on this composed entity? Indeed, coincidence measurements are global measurements on the composed entity. This is exactly what we showed in \cite{aertssozzo2014}, namely, the entanglement contained in the data we collected from the coincidence experiments on the composed entity {\it The Animal Acts} cannot be modeled in such a way that it is all concentrated in the state of the composed entity while the coincidence measurements themselves would be represented by product self-adjoint operators. And here again, in both web tests, we find a solution in which both the pre-measurement state and the coincidence measurements are entangled. This shows the systematic presence of quantum structures in the way concepts combine on the web and an even stronger form of entanglement than the one typically identified in quantum physics, where it is believed that the entanglement can only be concentrated in the state of the composed entity.

For the sake of completeness, we summarize the content of this paper in the following.

In Sect. \ref{data}, we describe the results of the web tests on the corpuses of documents `Google Books' and COCA, where in the latter we used the `collocates technique' (`Collocates test'). We observe that the CHSH inequality is manifestly violated, thus suggesting that entanglement is present in the conceptual combination {\it The Animal Acts} and possibly also in the coincidence measurements themselves. In Sect. \ref{quantumrep}, we provide a faithful mathematical representation in complex Hilbert space of the data presented in Sect. \ref{data}, finding some interesting results that confirm that quantum entanglement represents genuine meaning connections between sentences and concepts. We finally offer in Sect. \ref{conclusion} some conclusive remarks.

\section{Violation of the CHSH inequality in corpuses of documents\label{data}}
Let us start this section by explicitly writing the CHSH version of Bell's inequality \cite{clauser1969}
\begin{equation} \label{chsh}
-2 \le E(A',B')+E(A',B)+E(A,B')-E(A,B) \le 2
\end{equation}
Like we did in \cite{aertssozzo2011} when we realized experimentally a violation of (\ref{chsh}) on the conceptual entity {\it The Animal Acts} which is the combination of two conceptual entities {\it Animal} and {\it Acts}, we look at this combination as a `conceptual version' of how the two spin quantum entities are `combined' intouyghjuj the quantum entity consisting of the both spin quantum entities in a singlet spin state. This is in line with the way we have applied the quantum formalism to model concepts and their combinations within our approach to quantum cognition \cite{aerts2009,aertsbroekaertgaborasozzo2013,aertsgaborasozzo2013,aertsIJTP,aertsetal2018e}. It is also in line with this `conceptual version of combination' which we have presented in \cite{aertssozzo2014}, where we represented 
the data within a quantum model in complex Hilbert space, proving that a strong form of `entanglement' exists between the component concepts {\it Animal} and {\it Acts}. This is what we will do in the present paper with the data collected in the corpuses of documents.

Let us first summarize in this section the results we obtained in two types of a web test on the same conceptual combination, where data on {\it The Animal Acts} were collected in searches on corpuses of documents, specifically, the corpuses `Google Books' and COCA  \cite{beltrangeriente2018} (see Sect. \ref{intro}). We consider the 
sentence {\it The Animal Acts} as a combination of the concepts {\it Animal} and {\it Acts}. Then, we consider two pairs of exemplars, or states, of {\it Animal}, namely ({\it Horse}, {\it Bear}) and ({\it Tiger}, {\it Cat}), and two pairs of exemplars, or states, of {\it Acts}, namely ({\it Growls}, {\it Whinnies}) and ({\it Snorts}, {\it Meows}).

In the cognitive test in \cite{aertssozzo2011}, the coincidence experiments $AB$, $AB'$, $A'B$ and $A'B'$ consisted in asking a sample of 81 human participants to answer the question `is a good example of' the concept {\it The Animal Acts}. In experiment $AB$, participants chose among the four possibilities (1) {\it The Horse Growls}, (2) {\it The Bear Whinnies}  -- in which cases an outcome $+1$ was registered -- and (3) {\it The Horse Whinnies}, (4) {\it The Bear Growls} -- in which cases an outcome $-1$ was registered. Analogously, in experiment $AB'$, participants chose among the four possibilities (1) {\it The Horse Snorts}, (2) {\it The Bear Meows}  -- in which cases an outcome $+1$ was registered -- and (3) {\it The Horse Meows}, (4) {\it The Bear Snorts} -- in which cases an outcome $-1$ was registered. Next, in experiment $A'B$, participants chose among the four possibilities (1) {\it The Tiger Growls}, (2) {\it The Cat Whinnies}  -- in which cases an outcome $+1$ was registered -- and (3) {\it The Tiger Whinnies}, (4) {\it The Cat Growls} -- in which cases an outcome $-1$ was registered. Finally, in experiment $A'B'$, participants chose among the four possibilities (1) {\it The Tiger Snorts}, (2) {\it The Cat Meows}  -- in which cases an outcome $+1$ was registered -- and (3) {\it The Tiger Meows}, (4) {\it The Cat Snorts} -- in which cases an outcome $-1$ was registered. We computed the judged probabilities as large number limits of relative frequencies of responses, inserting them into the expectation values appearing in the  CHSH inequality (\ref{chsh}) and found a value of 2.4197, hence a violation, which was statistically significant \cite{aertssozzo2011}.

In the `Google Books test' in \cite{beltrangeriente2018}, the coincidence operation $AB$ consisted in retrieving from `Google Books' the frequencies of appearance of the four strings  ``horse growls'', ``horse whinnies'', ``bear growls'' and ``bear whinnies''. We respectively found the following number of entries 0, 464, 247 and 0. Thus, the four strings overall appeared 711 times. We then computed the relative frequency of appearance by dividing the number of appearance of each string by the total number of appearance. The four relative frequencies for ``horse growls'', ``horse whinnies'', ``bear growls'' and ``bear whinnies'' constitute an estimation of the `probabilities of appearance' $\mu(HG)$, $\mu(HW)$, $\mu(BG)$ and $\mu(BW)$, respectively. Data are reported in Table 1, coincidence operation $AB$. Next, we attached the outcome $+1$ to the searches ``horse growls'' and ``bear whinnies'', and the outcome $-1$ to the searches ``horse whinnies'' and ``bear growls'' and calculated the expectation value
\begin{equation} \label{GB_AB}
E(A,B)=\mu(HG)-\mu(HW)-\mu(BG)+\mu(BW)=-1
\end{equation}
The coincidence operation $AB'$ consisted in retrieving from `Google Books' the frequencies of appearance of the four strings  ``horse snorts'', ``horse meows'', ``bear snorts'' and ``bear meows''. We respectively found the following number of entries 202, 0, 0 and 0, thus the four strings overall appeared 202 times. We then computed the relative frequency of appearance of ``horse snorts'', ``horse meows'', ``bear snorts'' and ``bear meows'', hence the corresponding probabilities of appearance $\mu(HS)$, $\mu(HM)$, $\mu(BS)$ and $\mu(BM)$, respectively. Data are reported in Table 1, coincidence operation $AB'$. Next, we attached the outcome $+1$ to the searches ``horse snorts'' and ``bear meows'', and the outcome $-1$ to the searches ``horse meows'' and ``bear snorts'' and calculated the expectation value
\begin{equation} \label{GB_AB'}
E(A,B')=\mu(HS)-\mu(HM)-\mu(BS)+\mu(BM)=1
\end{equation}
The coincidence operation $A'B$ consisted in retrieving from `Google Books' the frequencies of appearance of the four strings  ``tiger growls'', ``tiger whinnies'', ``cat growls'' and ``cat whinnies''. We respectively found the following number of entries 97, 0, 41 and 0, thus the four strings overall appeared 138 times. We then computed the relative frequency of appearance of ``tiger growls'', ``tiger whinnies'', ``cat growls'' and ``cat whinnies'', hence the corresponding probabilities of appearance $\mu(TG)$, $\mu(TW)$, $\mu(CG)$ and $\mu(CW)$, respectively. Data are reported in Table 1, coincidence operation $A'B$. Next, we attached the outcome $+1$ to the searches ``tiger growls'' and ``cat whinnies'', and the outcome $-1$ to the searches ``tiger whinnies'' and ``cat growls'' and calculated the expectation value
\begin{equation} \label{GB_A'B}
E(A',B)=\mu(TG)-\mu(TW)-\mu(CG)+\mu(CW)=0.4058
\end{equation}
The coincidence operation $A'B'$ consisted in retrieving from `Google Books' the frequencies of appearance of the four strings  ``tiger snorts'', ``tiger meows'', ``cat snorts'' and ``cat meows''. We respectively found the following number of entries 0, 0, 0 and 331, thus the four strings overall appeared 331 times. We then computed the relative frequency of appearance of ``tiger snorts'', ``tiger meows'', ``cat snorts'' and ``cat meows'', hence the corresponding probabilities of appearance $\mu(TS)$, $\mu(TM)$, $\mu(CS)$ and $\mu(CM)$, respectively. Data are reported in Table 1, coincidence operation $A'B'$. Next, we attached the outcome $+1$ to the searches ``tiger snorts'' and ``cat meows'', and the outcome $-1$ to the searches ``tiger meows'' and ``cat snorts'' and calculated the expectation value
\begin{equation} \label{GB_A'B'}
E(A',B')=\mu(TS)-\mu(TM)-\mu(CS)+\mu(CM)=1
\end{equation}
Using the values in (\ref{GB_AB})--(\ref{GB_A'B'}), the intermediate term in the CHSH inequality (\ref{chsh}) becomes
\begin{equation}
E(A',B')+E(A',B)+E(A,B')-E(A,B)=3.41
\end{equation}
Hence, the CHSH inequality is strongly violated and the amount of violation exceed the known `Cirel'son bound' $2\sqrt{2}$, which holds for entangled states and product measurements on quantum entities \cite{cirelson1980}. 
%SS (see Sect. \ref{operational}).
\begin{table} \label{tab1}
\centering
\begin{footnotesize}
\begin{tabular}{|c |c | c | c| c| }
\hline
Operation \textrm{$AB$} & ``horse growls'' & ``horse whinnies'' & ``bear growls'' & ``bear whinnies'' \\
 & $\mu(HG)=0$ & $\mu(HW)=0.6526$ & $\mu(BG)=0.3474$ & $\mu(BW)=0$  \\
\hline
Operation \textrm{$AB'$} & ``horse snorts'' & ``horse meows'' & ``bear snorts'' & ``bear meows'' \\
 & $\mu(HS)=1$ & $\mu(HM)=0$ & $\mu(BS)=0$ & $\mu(BM)=0$  \\
\hline
Operation \textrm{$A'B$} & ``tiger growls'' & ``tiger whinnies'' & ``cat growls'' & ``cat whinnies'' \\
 & $\mu(TG)=0.7029$ & $\mu(TW)=0$ & $\mu(CG)=0.2971$ & $\mu(CW)=0$  \\
\hline
Operation \textrm{$A'B'$} & ``tiger snorts'' & ``tiger meows'' & ``cat snorts'' & ``cat meows'' \\
 & $\mu(TS)=0$ & $\mu(TM)=0$ & $\mu(CS)=0$ & $\mu(CM)=1$  \\
\hline
\end{tabular}
\caption{The data collected in the coincidence operations on entanglement in the web test using `Google Books' as corpus of documents \cite{beltrangeriente2018}.}
\end{footnotesize}
\end{table}
In \cite{beltrangeriente2018}, we also used other corpuses of documents, like the corpuses NOW and COCA (see Sect. \ref{intro}), finding very similar results with respect to the violation of the CHSH inequality, namely, a violation of 3 in NOW and a violation of 3.33 in COCA.

In all the web tests above, searches were made in corpuses of text looking for exact strings of characters, i.e. a search for the frequency of appearance of ``horse whinnies'' was a search for the frequency of appearance of the exact string of characters contained in ``horse whinnies''. A less sharp way of identifying meaning connections in corpuses of texts uses the technique of `collocates'. Consider, for example, the words ``horse'' and ``whinnies''. In a collocates technique search, one of these words, say ``horse'', is the `target word' and it is considered as the center of an interval of words. One can indicate the number of words that the width of an interval with in its center the target word can have, and we chose for our search the maximum number available in COCA to be equal to  9 words. This means that, whenever the second word ``whinnies'' is spotted in a search in the interval of 19 words, 9 words to the left of ``horse'' and 9 words to its right, it will be registered as a co-occurrence of both words ``horse'' and  ``whinnies''. The aim of the collocates technique is to loosen the strictness of co-occurrence and already allow such a less strict co-occurrence to be counted in case the target word ``horse'' and the  collocate word ``whinnies'' appear in each others neighbourhood. The `collocates technique' seem to more closely resemble the way human mind combines concepts in practice.

Thus, we calculated in \cite{beltrangeriente2018} relative frequencies of co-occurrencies in the web test on COCA using the collocates technique, and we will call the `Collocates test' the new collection of probabilities of appearance. Using the same symbols above, we performed the coincidence operations $AB$, $AB'$, $A'B$ and $A'B'$ finding the results in Table 2.
\begin{table} \label{tab2}
\centering
\begin{footnotesize}
\begin{tabular}{|c |c | c | c| c| }
\hline
Operation \textrm{$AB$} & ``horse growls'' & ``horse whinnies'' & ``bear growls'' & ``bear whinnies'' \\
 & $\mu(HG)=0$ & $\mu(HW)=0.8$ & $\mu(BG)=0.2$ & $\mu(BW)=0$  \\
\hline
Operation \textrm{$AB'$} & ``horse snorts'' & ``horse meows'' & ``bear snorts'' & ``bear meows'' \\
 & $\mu(HS)=1$ & $\mu(HM)=0$ & $\mu(BS)=0$ & $\mu(BM)=0$  \\
\hline
Operation \textrm{$A'B$} & ``tiger growls'' & ``tiger whinnies'' & ``cat growls'' & ``cat whinnies'' \\
 & $\mu(TG)=0.4$ & $\mu(TW)=0$ & $\mu(CG)=0.6$ & $\mu(CW)=0$  \\
\hline
Operation \textrm{$A'B'$} & ``tiger snorts'' & ``tiger meows'' & ``cat snorts'' & ``cat meows'' \\
 & $\mu(TS)=0$ & $\mu(TM)=0$ & $\mu(CS)=0$ & $\mu(CM)=1$  \\
\hline
\end{tabular}
\caption{The data collected in the coincidence operations on entanglement in the web test using COCA as corpus of documents and applying the collocates technique \cite{beltrangeriente2018}.}
\end{footnotesize}
\end{table}
The corresponding expectation values are
\begin{eqnarray}
E(A,B)&=&\mu(HG)-\mu(HW)-\mu(BG)+\mu(BW)=-1 \\
E(A,B')&=&\mu(HS)-\mu(HM)-\mu(BS)+\mu(BM)=1 \\
E(A',B)&=&\mu(TG)-\mu(TW)-\mu(CG)+\mu(CW)=-0.2 \\
E(A',B')&=&\mu(TS)-\mu(TM)-\mu(CS)+\mu(CM)=1
\end{eqnarray}
for a value of the intermediate term of the CHSH inequality equal to
\begin{equation}
E(A',B')+E(A',B)+E(A,B')-E(A,B)=2.8
\end{equation}
This time the violation of the CHSH inequality has a numerical value within the `Cirel'son bound' and is closer to the CHSH violation in the cognitive test, as expected.

As we can see, in the web tests we have presented in this section, we found a systematic violation of the CHSH inequality. 
 The next section will show how this entanglement can be modeled within the Hilbert space formalism of quantum theory.

\section{Quantum modeling of data on corpuses of documents\label{quantumrep}}
We elaborate a quantum representation in Hilbert space of the web tests data in Sect. \ref{data} on the conceptual combination {\it The Animal Acts}, following the technical procedures developed in \cite{aertssozzo2014}. As mentioned in Sect. \ref{intro}, and shown in \cite{aertssozzo2014}, the way states are prepared and measurements are performed in {\it The Animal Acts} entails that, not only the state of the conceptual entity {\it The Animal Acts} is entangled, but also the four coincidence measurements are generally entangled. 

Let us start from the realistic-operational foundation of cognitive entities, where concepts and their combinations are described as entities in specific states which change under the effect of measurements \cite{aertssassolidebianchisozzo2016}. Then, let us consider the search operations described in Sect. \ref{data}. From a realistic-operational point of view, the initial situation of the composed conceptual entity {\it The Animal Acts}, before any operation is performed, can be described by a state $p$, which expresses the potentiality to actualize correlations when measurements are performed (cognitive test, web search operation, etc.). 

Experiment, or operation, $AB$ corresponds to a coincidence measurement $e_{AB}$ performed on the composite conceptual entity {\it The Animal Acts} with four possible outcomes $\lambda_{HG}$, $\lambda_{BW}$ (which are chosen equal to $+1$), $\lambda_{HW}$ and $\lambda_{BG}$ (which are chosen equal to $-1$), and four outcome states $p_{HG}$, $p_{BW}$, $p_{HW}$ and $p_{BG}$, describing the situation of {\it The Animal Acts} once (1), (2), (3) and (4), respectively, occur in $AB$. Let us denote by $p_{p}(HG)$, $p_{p}(BW)$, $p_{p}(HW)$ and $p_{p}(BG)$ the probability that the outcome $\lambda_{HG}$, $\lambda_{BW}$, $\lambda_{HW}$ and $\lambda_{BG}$, respectively, is obtained when {\it The Animal Acts} is in the state $p$ and the measurement $e_{AB}$ 
%%SS
is performed.\footnote{For $AB$, the appearance numbers in Tables 1 and 2, Sect. \ref{data}, are an estimation of the probabilities that a given outcome is obtained. The same remark applies to $AB'$, $A'B$ and $A'B'$.}

Experiment, or operation, $AB'$ corresponds to a coincidence measurement $e_{AB'}$ on the composite conceptual entity {\it The Animal Acts} with four possible outcomes $\lambda_{HS}$, $\lambda_{BM}$ (which are chosen equal to $+1$), $\lambda_{HS}$ and $\lambda_{BM}$ (which are chosen equal to $-1$), and four outcome states $p_{HS}$, $p_{BM}$, $p_{HM}$ and $p_{BS}$, describing the situation of {\it The Animal Acts} once (1), (2), (3) and (4), respectively, occur in $AB'$. Let us denote by $p_{p}(HS)$, $p_{p}(BM)$, $p_{p}(HM)$ and $p_{p}(BS)$ the probability that the outcome $\lambda_{HS}$, $\lambda_{BM}$, $\lambda_{HM}$ and $\lambda_{BS}$, respectively, is obtained when {\it The Animal Acts} is in the state $p$ and the measurement $e_{AB'}$ is performed.

Experiment, or operation, $A'B$ corresponds to a coincidence measurement $e_{A'B}$ on the composite conceptual entity {\it The Animal Acts} with four possible outcomes $\lambda_{TG}$, $\lambda_{CW}$ (which are chosen equal to $+1$), $\lambda_{TW}$ and $\lambda_{CG}$ (which are chosen equal to $-1$), and four outcome states $p_{TG}$, $p_{CW}$, $p_{TW}$ and $p_{CG}$, describing the situation of {\it The Animal Acts} once (1), (2), (3) and (4), respectively, occur in $A'B$. Let us denote by $p_{p}(TG)$, $p_{p}(CW)$, $p_{p}(TW)$ and $p_{p}(CG)$ the probability that the outcome $\lambda_{TG}$, $\lambda_{CW}$, $\lambda_{TW}$ and $\lambda_{CG}$, respectively, is obtained when {\it The Animal Acts} is in the state $p$ and the measurement $e_{A'B}$ is performed.

Experiment, or operation, $A'B'$ corresponds to a coincidence measurement $e_{A'B'}$ on the composite conceptual entity {\it The Animal Acts} with four possible outcomes $\lambda_{TS}$, $\lambda_{CM}$ (which are chosen equal to $+1$), $\lambda_{TM}$ and $\lambda_{CS}$ (which are chosen equal to $-1$), and four outcome states $p_{TS}$, $p_{CM}$, $p_{TM}$ and $p_{CS}$, describing the situation of {\it The Animal Acts} once (1), (2), (3) and (4), respectively, occur in $A'B'$. Let us denote by $p_{p}(TS)$, $p_{p}(CM)$, $p_{p}(TM)$ and $p_{p}(CS)$ the probability that the outcome $\lambda_{TS}$, $\lambda_{CM}$, $\lambda_{TM}$ and $\lambda_{CS}$, respectively, is obtained when {\it The Animal Acts} is in the state $p$ and the measurement $e_{A'B'}$ is performed.

Let us preliminarily observe that the four coincidence measurements $e_{AB}$, $e_{AB'}$, $e_{A'B}$ and $e_{A'B'}$ correspond to operations performed on the overall entity {\it The Animal Acts} rather than to operations separately performed on the individual entities {\it Animal} and {\it Acts}. Furthermore, their outcome states, for example, {\it Horse Whinnies} or {\it Tiger Growls}, are again combinations of concepts, hence they should in principle be described as entangled states, if entanglement has to mathematically captures the meaning connection existing between concepts. This concretely means that a coincidence measurement, say $e_{AB}$, on {\it The Animal Acts} cannot be generally decomposed into a sub measurement on {\it Animal} and a sub measurement on {\it Acts}, though, e.g., {\it Horse Growls} is syntactically formed by juxtaposing the words ``horse'' and ``growls'', because the concepts {\it Horse} and {\it Growls} are connected by meaning.

This remark will become evident in the following mathematical representation.
%% D
Let us consider the combined entity {\it The Animal Acts} and let us associate it with the Hilbert space $\mathbb{C}^4$ of all ordered 4-tuples of complex numbers, which is canonically isomorphic to the tensor product Hilbert space $\mathbb{C}^2 \otimes \mathbb{C}^2$, where $\mathbb{C}^2$ is the complex Hilbert space of all ordered 2-tuples of complex numbers. Let $\{(1,0,0,0),(0,1,0,0),(0,0,1,0),(0,0,0,1)\}$ be the canonical ON basis of $\mathbb{C}^4$. 
We consider the isomorphism where this basis corresponds to the ON basis of $\mathbb{C}^2 \otimes \mathbb{C}^2$ made up by the unit vectors $(1,0)\otimes (1,0)$, $(1,0)\otimes (0,1)$, $(0,1)\otimes (1,0)$ and $(0,1)\otimes (0,1)$. In the canonical ON basis of $\mathbb{C}^4$, the initial state $p$ of {\it The Animal Acts} is represented by the unit vector $|p\rangle=(ae^{i \alpha}, be^{i \beta}, ce^{i \gamma}, de^{i \delta})$, where $a,b,c,d$ are non-negative real numbers such that $a^2+b^2+c^2+d^2=1$ and $\alpha$, $\beta$, $\gamma$, $\delta$ are real numbers.
%%SS
It is known that $|p\rangle$ represents a product state if and only if %%SS $ade^{i(\alpha+\delta)}+bce^{i(\beta+\gamma)}=0$
$ade^{i(\alpha+\delta)}-bce^{i(\beta+\gamma)}=0$, otherwise $p$ represents an entangled state.

Let us come to the representation of the coincidence measurements. The coincidence measurement $e_{AB}$ has four outcomes $\lambda_{HG}$, $\lambda_{HW}$, $\lambda_{BG}$ and $\lambda_{BW}$, corresponding to {\it Horse Growls}, {\it Bear Whinnies}, {\it Bear Growls} and {\it Bear Whinnies}, and four eigenstates $p_{HG}$, $p_{HW}$, $p_{BG}$ and $p_{BW}$, respectively. Let us represent the measurement $e_{AB}$ by the self-adjoint operator ${\mathscr E}_{AB}$ or, equivalently, by the spectral family $\{|p_{HG}\rangle\langle p_{HG}|, |p_{HW}\rangle\langle p_{HW}|, |p_{BG}\rangle\langle p_{BG}|, |p_{BW}\rangle\langle p_{BW}| \}$, such that the eigenstates $p_{HG}$, $p_{HW}$, $p_{BG}$ and $p_{BW}$ are respectively represented by the eigenvectors of ${\mathscr E}_{AB}$
\begin{eqnarray}
|p_{HG}\rangle&=&(a_{HG}e^{i \alpha_{HG}}, b_{HG}e^{i \beta_{HG}}, c_{HG}e^{i \gamma_{HG}}, d_{HG}e^{i \delta_{HG}}) \label{HG}\\
|p_{HW}\rangle&=&(a_{HW}e^{i \alpha_{HW}}, b_{HW}e^{i \beta_{HW}}, c_{HW}e^{i \gamma_{HW}}, d_{HW}e^{i \delta_{HW}}) \label{HW}\\
|p_{BG}\rangle&=&(a_{BG}e^{i \alpha_{BG}}, b_{BG}e^{i \beta_{BG}}, c_{BG}e^{i \gamma_{BG}}, d_{BG}e^{i \delta_{BG}}) \label{BG}\\
|p_{BW}\rangle&=&(a_{BW}e^{i \alpha_{BW}}, b_{BW}e^{i \beta_{BW}}, c_{BW}e^{i \gamma_{BW}}, d_{BW}e^{i \delta_{BW}}) \label{BW}
\end{eqnarray}
In (\ref{HG})--(\ref{BW}), $a_{ij}, b_{ij}, c_{ij}, d_{ij}$ are non-negative real numbers and $\alpha_{ij}, \beta_{ij},\gamma_{ij}, \delta_{ij}$ are real numbers, $i=H,B; j=G,W$. The self-adjoint operator ${\mathscr E}_{AB}$ can be expressed as a tensor product operator if and only if all unit vectors in (\ref{HG})--(\ref{BW}) represent product states, otherwise ${\mathscr E}_{AB}$ is entangled, hence $e_{AB}$ is an entangled measurement.
 
Now, to determine the unit vectors in (\ref{HG})--(\ref{BW}) that satisfy empirical data, we need to impose 
%%SS 
specific conditions, as follows.

(i) Normalization. The vectors in (\ref{HG})--(\ref{BW})  have to be unitary:
\begin{eqnarray}
a_{HG}^2+b_{HG}^2+c_{HG}^2+d_{HG}^2&=&1 \nonumber  \\
a_{HW}^2+b_{HW}^2+c_{HW}^2+d_{HW}^2&=&1 \nonumber \\
a_{BG}^2+b_{BG}^2+c_{BG}^2+d_{BG}^2&=&1 \nonumber \\
a_{BW}^2+b_{BW}^2+c_{BW}^2+d_{BW}^2&=&1 \nonumber
\end{eqnarray}

(ii) Orthogonality. The vectors in (\ref{HG})--(\ref{BW})  have to be mutually orthogonal:
\begin{eqnarray}
0=\langle p_{HG}|p_{HW} \rangle&=&a_{HG}a_{HW}e^{i(\alpha_{HW}-\alpha_{HG})}+b_{HG}b_{HW}e^{i(\beta_{HW}-\beta_{HG})}+ \nonumber \\ &+&c_{HG}c_{HW}c^{i(\gamma_{HW}-\gamma_{HG})}+d_{HG}d_{HW}e^{i(\delta_{HW}-\delta_{HG})} \nonumber \\
0=\langle p_{HG}|p_{BG} \rangle&=&a_{HG}a_{BG}e^{i(\alpha_{BG}-\alpha_{HG})}+b_{HG}b_{BG}e^{i(\beta_{BG}-\beta_{HG})}+ \nonumber \\  &+&c_{HG}c_{BG}c^{i(\gamma_{BG}-\gamma_{HG})}+d_{HG}d_{BG}e^{i(\delta_{BG}-\delta_{HG})} \nonumber \\
0=\langle p_{HG}|p_{BW} \rangle&=&a_{HG}a_{BW}e^{i(\alpha_{BW}-\alpha_{HG})}+b_{HG}b_{BW}e^{i(\beta_{BW}-\beta_{HG})}+ \nonumber \\ &+&c_{HG}c_{BW}c^{i(\gamma_{BW}-\gamma_{HG})}+d_{HG}d_{BW}e^{i(\delta_{BW}-\delta_{HG})} \nonumber \\
0=\langle p_{HW}|p_{BG} \rangle&=&a_{HW}a_{BG}e^{i(\alpha_{BG}-\alpha_{HW})}+b_{HW}b_{BG}e^{i(\beta_{BG}-\beta_{HW})}+ \nonumber \\ &+&c_{HW}c_{BG}c^{i(\gamma_{BG}-\gamma_{HW})}+d_{HW}d_{BG}e^{i(\delta_{BG}-\delta_{HW})} \nonumber \\
0=\langle p_{HW}|p_{BW} \rangle&=&a_{HW}a_{BW}e^{i(\alpha_{BW}-\alpha_{HW})}+b_{HW}b_{BW}e^{i(\beta_{BW}-\beta_{HW})}+ \nonumber \\ &+&c_{HW}c_{BW}c^{i(\gamma_{BW}-\gamma_{HW})}+d_{HW}d_{BW}e^{i(\delta_{BW}-\delta_{HW})} \nonumber \\
0=\langle p_{BG}|p_{BW} \rangle&=&a_{BG}a_{BW}e^{i(\alpha_{BW}-\alpha_{BG})}+b_{BG}b_{BW}e^{i(\beta_{BW}-\beta_{BG})}+ \nonumber \\ &+&c_{BG}c_{BW}c^{i(\gamma_{BW}-\gamma_{BG})}+d_{BG}d_{BW}e^{i(\delta_{BW}-\delta_{BG})} \nonumber
\end{eqnarray}

(iii) %%SS Experimental probability
Data representation. Let us consider the probabilities of appearance $\mu(HG)$, $\mu(HW)$, $\mu(BG)$ and $\mu(BW)$ for {\it The Horse Growls}, {\it The Bear Whinnies}, {\it The Bear Growls} and {\it Bear Whinnies}, respectively (reported in Tables 1 and 2). Because of the Born rule of quantum probability, we have:
\begin{eqnarray}
\mu(HG)=|\langle p_{HG}|p\rangle|^{2}&=&a^2 a_{HG}^2+b^2 b_{HG}^2+c^2 c_{HG}^2+d^2 d_{HG}^2+ \nonumber \\
&+&2aba_{HG}b_{HG}\cos(\alpha-\alpha_{HG}-\beta+\beta_{HG})+ \nonumber \\
&+&2aca_{HG}c_{HG}\cos(\alpha-\alpha_{HG}-\gamma+\gamma_{HG})+ \nonumber \\
&+&2ada_{HG}d_{HG}\cos(\alpha-\alpha_{HG}-\delta+\delta_{HG})+ \nonumber \\
&+&2bcb_{HG}c_{HG}\cos(\beta-\beta_{HG}-\gamma+\gamma_{HG})+ \nonumber \\
&+&2bdb_{HG}d_{HG}\cos(\beta-\beta_{HG}-\delta+\delta_{HG})+ \nonumber \\
&+&2cdc_{HG}d_{HG}\cos(\gamma-\gamma_{HG}-\delta+\delta_{HG}) \nonumber
\end{eqnarray}
\begin{eqnarray}
\mu(HW)=|\langle p_{HW}|p\rangle|^{2}&=&a^2 a_{HW}^2+b^2 b_{HW}^2+c^2 c_{HW}^2+d^2 d_{HW}^2+ \nonumber \\
&+&2aba_{HW}b_{HW}\cos(\alpha-\alpha_{HW}-\beta+\beta_{HW})+ \nonumber \\
&+&2aca_{HW}c_{HW}\cos(\alpha-\alpha_{HW}-\gamma+\gamma_{HW})+ \nonumber \\
&+&2ada_{HW}d_{HW}\cos(\alpha-\alpha_{HW}-\delta+\delta_{HW})+ \nonumber \\
&+&2bcb_{HW}c_{HW}\cos(\beta-\beta_{HW}-\gamma+\gamma_{HW})+ \nonumber \\
&+&2bdb_{HW}d_{HW}\cos(\beta-\beta_{HW}-\delta+\delta_{HW})+ \nonumber \\
&+&2cdc_{HW}d_{HW}\cos(\gamma-\gamma_{HW}-\delta+\delta_{HW}) \nonumber
\end{eqnarray}
\begin{eqnarray}
\mu(BG)=|\langle p_{BG}|p\rangle|^{2}&=&a^2 a_{BG}^2+b^2 b_{BG}^2+c^2 c_{BG}^2+d^2 d_{BG}^2+ \nonumber \\
&+&2aba_{BG}b_{BG}\cos(\alpha-\alpha_{BG}-\beta+\beta_{BG})+ \nonumber \\
&+&2aca_{BG}c_{BG}\cos(\alpha-\alpha_{BG}-\gamma+\gamma_{BG})+ \nonumber \\
&+&2ada_{BG}d_{BG}\cos(\alpha-\alpha_{BG}-\delta+\delta_{BG})+ \nonumber \\
&+&2bcb_{BG}c_{BG}\cos(\beta-\beta_{BG}-\gamma+\gamma_{BG})+ \nonumber \\
&+&2bdb_{BG}d_{BG}\cos(\beta-\beta_{BG}-\delta+\delta_{BG})+ \nonumber \\
&+&2cdc_{BG}d_{BG}\cos(\gamma-\gamma_{BG}-\delta+\delta_{BG}) \nonumber
\end{eqnarray}
\begin{eqnarray}
\mu(BW)=|\langle p_{BW}|p\rangle|^{2}&=&a^2 a_{BW}^2+b^2 b_{BW}^2+c^2 c_{BW}^2+d^2 d_{BW}^2+ \nonumber \\
&+&2aba_{BW}b_{BW}\cos(\alpha-\alpha_{BW}-\beta+\beta_{BW})+ \nonumber \\
&+&2aca_{BW}c_{BW}\cos(\alpha-\alpha_{BW}-\gamma+\gamma_{BW})+ \nonumber \\
&+&2ada_{BW}d_{BW}\cos(\alpha-\alpha_{BW}-\delta+\delta_{BW})+ \nonumber \\
&+&2bcb_{BW}c_{BW}\cos(\beta-\beta_{BW}-\gamma+\gamma_{BW})+ \nonumber \\
&+&2bdb_{BW}d_{BW}\cos(\beta-\beta_{BW}-\delta+\delta_{BW})+ \nonumber \\
&+&2cdc_{BW}d_{BW}\cos(\gamma-\gamma_{BW}-\delta+\delta_{BW}) \nonumber
\end{eqnarray}

Before coming to an explicit solution of the equations above, let us consider the other coincidence measurements.

The coincidence measurement $e_{AB'}$ has four outcomes $\lambda_{HS}$, $\lambda_{HM}$, $\lambda_{BS}$ and $\lambda_{BM}$, corresponding to {\it Horse Snorts}, {\it Horse Meows}, {\it Bear Snorts} and {\it Bear Meows}, and four eigenstates $p_{HS}$, $p_{HM}$, $p_{BS}$ and $p_{BM}$, respectively. The measurement $e_{AB'}$ is represented by the self-adjoint operator ${\mathscr E}_{AB'}$ or, equivalently, by the spectral family $\{|p_{HS}\rangle\langle p_{HS}|, |p_{HM}\rangle\langle p_{HM}|, |p_{BS}\rangle\langle p_{BS}|, |p_{BM}\rangle\langle p_{BM}| \}$, such that the eigenstates $p_{HS}$, $p_{HM}$, $p_{BS}$ and $p_{BM}$ are represented by eigenvectors whose form is identical to (\ref{HG})--(\ref{BW}) satisfying conditions (i)--(iii), with obvious symbol replacement.

The coincidence measurement $e_{A'B}$ has four outcomes $\lambda_{TG}$, $\lambda_{TW}$, $\lambda_{CG}$ and $\lambda_{CW}$, corresponding to {\it Tiger Growls}, {\it Tiger Whinnies}, {\it Cat Growls} and {\it Cat Whinnies}, and four eigenstates $p_{TG}$, $p_{TW}$, $p_{CG}$ and $p_{CW}$, respectively. The  measurement $e_{A'B}$ is represented by the self-adjoint operator ${\mathscr E}_{A'B}$ or, equivalently, by the spectral family $\{|p_{TG}\rangle\langle p_{TG}|, |p_{TW}\rangle\langle p_{TW}|, |p_{CG}\rangle\langle p_{CG}|, |p_{CW}\rangle\langle p_{CW}| \}$, such that the eigenstates $p_{TG}$, $p_{TW}$, $p_{CG}$ and $p_{CW}$ are represented by eigenvectors whose form is identical to (\ref{HG})--(\ref{BW}) satisfying conditions (i)--(iii), with obvious symbol replacement.

The coincidence measurement $e_{A'B'}$ has four outcomes $\lambda_{TS}$, $\lambda_{TM}$, $\lambda_{CS}$ and $\lambda_{CM}$, corresponding to {\it Tiger Snorts}, {\it Tiger Meows}, {\it Cat Snorts} and {\it Meows}, and four eigenstates $p_{TS}$, $p_{TM}$, $p_{CS}$ and $p_{CM}$, respectively. The measurement $e_{A'B'}$ is represented by the self-adjoint operator ${\mathscr E}_{A'B'}$ or, equivalently, by the spectral family $\{|p_{TS}\rangle\langle p_{TS}|, |p_{TM}\rangle\langle p_{TM}|, |p_{CS}\rangle\langle p_{CS}|, |p_{CM}\rangle\langle p_{CM}| \}$, such that the eigenstates $p_{TS}$, $p_{TM}$, $p_{CS}$ and $p_{CM}$ are represented by eigenvectors whose form is identical to (\ref{HG})--(\ref{BW}) satisfying conditions (i)--(iii), with obvious symbol replacement.

It is then sufficient to show how we can construct the self-adjoint operator ${\mathscr E}_{AB}$. The remaining operators are constructed in a similar way.

Let us preliminarily observe that the set (i)--(iii) identifies 20 equations which should be satisfied by 32 variables. To simplify the calculation, let us look for solutions such that:
\begin{eqnarray}
&&\alpha_{HG}=\beta_{HG}=\gamma_{HG}=\delta_{HG}=\Theta_{HG} \nonumber \\
&&\alpha_{HW}=\beta_{HW}=\gamma_{HW}=\delta_{HW}=\Theta_{HW}   \nonumber \\
&&\alpha_{BG}=\beta_{BG}=\gamma_{BG}=\delta_{BG}=\Theta_{BG}   \nonumber \\
&&\alpha_{BW}=\beta_{BW}=\gamma_{BW}=\delta_{BW}=\Theta_{BW}   \nonumber
\end{eqnarray}
In this way, (\ref{HG})--(\ref{BW}) become:
\begin{eqnarray}
|p_{HG}\rangle &=&e^{i \Theta_{HG}}(a_{HG}, b_{HG}, c_{HG}, d_{HG}) \\
|p_{HW}\rangle &=&e^{i \Theta_{HW}}(a_{HW}, b_{HW}, c_{HW}, d_{HW}) \\
|p_{BG}\rangle &=& e^{i \Theta_{BG}}(a_{BG}, b_{BG}, c_{BG}, d_{BG}) \\
|p_{BW}\rangle &=& e^{i \Theta_{BW}}(a_{BW}, b_{BW}, c_{BW}, d_{BW})
\end{eqnarray}

Then, let us represent represent the initial state of the combined concept {\it The Animal Acts} by the unit vector
\begin{equation}
|p\rangle=\frac{1}{\sqrt{2}}(0,1,-1,0) \label{singlet}
\end{equation}
The choice of this maximally entangled state is primarily motivated by the fact that this state leads to a solution with a simple and interesting interpretation, like we will see when we construct the solution. After constructing the solution, we will analyse more in detail 
%% the motivation for this choice.
the nature of the solutions and show how they represent well what we also intuitively know about meaning connection between concepts, taking into account our view that entanglement is linked to meaning connection.

Finally, let us set the outcomes $\lambda_{HG}=\lambda_{BW}=1$ and $\lambda_{HW}=\lambda_{BG}=-1$, as it is usual when dealing with correlation functions.

Let us firstly consider the `Google Books test'.

The solutions we find for the coincidence measurement $e_{AB}$ are the following.
\begin{eqnarray}
|p_{HG}\rangle =(0,0,0,-1) &\qquad& \Theta_{HG}=86.76^{\circ} \label{Gsol_HG} \\
|p_{HW}\rangle =(0,0.15,-0.99,0) &\qquad& \Theta_{HW}=71.75^{\circ} \label{Gsol_HW} \\
|p_{BG}\rangle =(0,0.99,0.15,0) &\qquad& \Theta_{BG}=0.14^{\circ} \label{Gsol_BG} \\
|p_{BW}\rangle =(1,0,0,0) &\qquad& \Theta_{BW}=21.69^{\circ} \label{Gsol_BW}
\end{eqnarray}
As we can see, the unit vectors $|p_{HG}\rangle$ and $|p_{BW}\rangle$ represent product states, while the unit vectors $|p_{HW}\rangle$ and $|p_{BG}\rangle$ represent entangled states. Hence, $e_{AB}$ is an entangled measurement,  containing however two eigenstates that are product states.

Analogously, we can proceed to find solutions for the coincidence measurement $e_{AB'}$, which are the following.
\begin{eqnarray}
|p_{HS}\rangle =(0,0.707,-0.707,0) &\qquad& \Theta_{HS}=0.11^{\circ} \label{Gsol_HS} \\
|p_{HM}\rangle =(0,0,0,-1) &\qquad& \Theta_{HM}=49.09^{\circ} \label{Gsol_HM} \\
|p_{BS}\rangle =(0,0.707,0.707,0) &\qquad& \Theta_{BS}=0.57^{\circ} \label{Gsol_BS} \\
|p_{BM}\rangle =(1,0,0,0) &\qquad& \Theta_{BM}=21.65^{\circ} \label{Gsol_BM}
\end{eqnarray}
As we can see, the unit vectors $|p_{HM}\rangle$ and $|p_{BM}\rangle$ represent product states, while the unit vectors $|p_{HS}\rangle$ and $|p_{BS}\rangle$ represent entangled states. Hence, $e_{AB'}$ is an entangled measurement,
containing again two eigenstates that are product states.

The solutions for the coincidence measurement $e_{A'B}$ are instead the following.
\begin{eqnarray}
|p_{TG}\rangle =(0,0.21,-0.98,0) &\qquad& \Theta_{TG}=71.75^{\circ} \label{Gsol_TG} \\
|p_{TW}\rangle =(0,0,0,-1) &\qquad& \Theta_{TW}=86.76^{\circ} \label{Gsol_TW} \\
|p_{CG}\rangle =(0,0.98,0.21,0) &\qquad& \Theta_{CG}=0.14^{\circ} \label{Gsol_CG} \\
|p_{CW}\rangle =(1,0,0,0) &\qquad& \Theta_{CW}=21.69^{\circ} \label{Gsol_CW}
\end{eqnarray}
As we can see, the unit vectors $|p_{TW}\rangle$ and $|p_{CW}\rangle$ represent product states, while the unit vectors $|p_{TG}\rangle$ and $|p_{CW}\rangle$ represent entangled states. Hence, $e_{A'B}$ is an entangled measurement,
containing again two eigenstates that are product states.

Finally, the solutions for the coincidence measurement $e_{A'B'}$ are the following.
\begin{eqnarray}
|p_{TS}\rangle =(1,0,0,0) &\qquad& \Theta_{TS}=21.65^{\circ} \label{Gsol_TS} \\
|p_{TM}\rangle =(0,0,-1,0) &\qquad& \Theta_{TM}=49.09^{\circ} \label{Gsol_TM} \\
|p_{CS}\rangle =(0,0.707,0.707,0) &\qquad& \Theta_{CS}=0.57^{\circ} \label{Gsol_CS} \\
|p_{CM}\rangle =(0,0.707,-0.707,0) &\qquad& \Theta_{CM}=0.11^{\circ} \label{Gsol_CM}
\end{eqnarray}
As we can see, the unit vectors $|p_{TS}\rangle$ and $|p_{TM}\rangle$ represent product states, while the unit vectors $|p_{CS}\rangle$ and $|p_{CM}\rangle$ represent entangled states. Hence, $e_{A'B'}$ is an entangled measurement,
containing again two eigenstates that are product states.

Next, let us present the solutions for the `Collocates test'. We have:
\begin{eqnarray}
|p_{HG}\rangle =(0,0,0,-1) &\qquad& \Theta_{HG}=86.76^{\circ} \label{Csol_HG} \\
|p_{HW}\rangle =(0,0.32,-0.95,0) &\qquad& \Theta_{HW}=71.84^{\circ} \label{Csol_HW} \\
|p_{BG}\rangle =(0,0.95,0.32,0) &\qquad& \Theta_{BG}=0.14^{\circ} \label{Csol_BG} \\
|p_{BW}\rangle =(1,0,0,0) &\qquad& \Theta_{BW}=21.69^{\circ} \label{Csol_BW} \\
|p_{HS}\rangle =(0,0.707,-0.707,0) &\qquad& \Theta_{HS}=0.11^{\circ} \label{Csol_HS} \\
|p_{HM}\rangle =(0,0,0,-1) &\qquad& \Theta_{HM}=49.09^{\circ} \label{Csol_HM} \\
|p_{BS}\rangle =(0,0.707,0.707,0) &\qquad& \Theta_{BS}=0.57^{\circ} \label{Csol_BS} \\
|p_{BM}\rangle =(1,0,0,0) &\qquad& \Theta_{BM}=21.65^{\circ} \label{Csol_BM}\\
|p_{TG}\rangle =(0,0.99,0.10,0) &\qquad& \Theta_{TG}=0.14^{\circ} \label{Csol_TG} \\
|p_{TW}\rangle =(0,0,0,-1) &\qquad& \Theta_{TW}=86.72^{\circ} \label{Csol_TW} \\
|p_{CG}\rangle =(0,0.10,-0.99,0) &\qquad& \Theta_{CG}=71.84^{\circ} \label{Csol_CG} \\
|p_{CW}\rangle =(1,0,0,0) &\qquad& \Theta_{CW}=21.69^{\circ} \label{Csol_CW} \\
|p_{TS}\rangle =(1,0,0,0) &\qquad& \Theta_{TS}=21.65^{\circ} \label{Csol_TS} \\
|p_{TM}\rangle =(0,0,-1,0) &\qquad& \Theta_{TM}=49.09^{\circ} \label{Csol_TM} \\
|p_{CS}\rangle =(0,0.707,0.707,0) &\qquad& \Theta_{CS}=0.57^{\circ} \label{Csol_CS} \\
|p_{CM}\rangle =(0,0.707,-0.707,0) &\qquad& \Theta_{CM}=0.11^{\circ} \label{Csol_CM}
\end{eqnarray}
As we can see, in each spectral family, we have two orthogonal projection operators on product state vectors and two orthogonal projection operators on entangled state vectors. Hence, the four coincidence measurements $e_{AB}$, $e_{AB'}$, $e_{A'B}$ and $e_{A'B'}$ are entangled measurements also in the case of the `Collocates test',
%% D
containing however each of them two product states
%% D
as eigenstates.

Some considerations on the quantum structures arising in the representation above should be made at this stage.

(1) In both web tests, all coincidence measurements are entangled. In addition, each coincidence measurement has two product eigenstates and two entangled eigenstates.

(2) In both web tests, all entangled eigenstates have the general representation $e^{i \Theta}(0,B,C,0)$, where $B,C, \Theta \in \Re$, $B^2+C^2=1$. This very symmetric and regular mathematical form derives from the choice of the maximally entangled represented by the unit vector $\frac{1}{\sqrt{2}}(0,1,-1,0)$ as pre-measurement state.

(3) In both web tests, the entangled eigenstates generally correspond to the exemplars of {\it The Animal Acts} that have a relatively higher meaning connection, e.g., the exemplars that trigger a relatively higher rate of responses, like {\it Horse Whinnies}, {\it Horse Snorts}, {\it Tiger Growls} and {\it Cat Meows}.

(4) In both web tests, the product eigenstates generally correspond to the exemplars of 
{\it The Animal Acts} that have a relatively low meaning connection, e.g., the exemplars that trigger a relatively low, or possibly null, rate of responses, like {\it Horse Meows}, {\it Bear Meows}, {\it Tiger Whinnies} and {\it Cat Whinnies}.

These remarks are important in our opinion, because they exactly express the fact that the quantum structure of entanglement captures both mathematically and conceptually meaning connection between sentences and concepts.

%%SS 
To conclude, we have proved in this section that the data collected in \cite{beltrangeriente2018} on corpuses of documents exhibit the same type of strong entanglement that was identified in the psychological test in \cite{aertssozzo2011}, namely, violation of the CHSH inequality is due to an entanglement that is present in both the initial state of the combined conceptual entity {\it The Animal Acts} and the coincidence measurements that are performed, due to the meaning connections between the component conceptual entities {\it Animal} and {\it Acts}. These results are 
%%SS port 
part 
of a more general theoretical framework on entanglement which we are currently developing \cite{aertsetal2019}.

\section{Conclusion\label{conclusion}}
We have worked out a quantum-theoretic model in complex Hilbert space for the combination of the two concepts {\it Animal} and {\it Acts} in the sentence {\it The Animal Acts}, and the data we collected on corpuses of documents for coincidence operations on this combination of concepts \cite{beltrangeriente2018}. The data have been shown  to violate CHSH inequalities and hence to identify the presence of entanglement \cite{beltrangeriente2018}. 
%%SS 
We have proved that, if we represent the state of the sentence {\it The Animal Acts} by means of a maximally entangled state (corresponding to the singlet spin state in quantum physics), and make use of the modeling scheme worked out in \cite{aertssozzo2014}, the coincidence measurements are entangled too, but such that the eigenstates corresponding to outcomes with negligible chance approach product states, while the eigenstates corresponding to probable outcomes are entangled. This confirms our view that entanglement expresses the meaning connection between concepts within a combination. Indeed, outcome states with low probability in the coincidence experiments carry low meaning connection within their combination while outcome states with high probability in the coincidence measurement carry high meaning content.

\section*{Acknowledgements}
This work was supported by QUARTZ (Quantum Information Access and Retrieval Theory), the Marie Sk{\l}odowska-Curie Innovative Training Network 721321 of the European Union's Horizon 2020 research and innovation programme.

\end{document}